\documentclass{article}

\pdfoutput=1

 \usepackage[nonatbib,final]{nips_2016}

\usepackage{nips_2016}
\usepackage[dvipsnames,usenames]{color}

\usepackage{graphicx}
\usepackage{amsmath,amssymb} 
\usepackage[utf8]{inputenc} 
\usepackage[T1]{fontenc}    
\usepackage{hyperref}       
\usepackage{url}            
\usepackage{booktabs}       
\usepackage{amsfonts}       
\usepackage{nicefrac}       
\usepackage{microtype}      

\title{MoCap-guided Data Augmentation \\for 3D Pose Estimation in the Wild} 

\author{
  Gr\'egory~Rogez \qquad Cordelia Schmid \\ 
\vspace{-3mm}
\\  
 Inria  Grenoble  Rh\^one-Alpes,  Laboratoire  Jean  Kuntzmann,  France  
\vspace{-3mm}
}

\begin{document}

\maketitle
\vspace{-2mm}

\begin{abstract}
\vspace{-3mm}

This paper addresses the problem of 3D human pose estimation 
in the wild. A significant challenge is the lack of training data,
i.e., 2D images of humans annotated with 3D poses. Such 
data is necessary to train state-of-the-art CNN architectures. Here, we propose
a solution to generate a large set of photorealistic synthetic images
of humans with 3D pose annotations. We introduce an image-based synthesis engine that
artificially augments a dataset of real images with 2D human pose annotations 
using 3D Motion Capture (MoCap) data.
Given a candidate 3D pose our algorithm selects for each joint an
image whose 2D pose locally matches the projected 3D pose. The
selected images are then combined to generate a new synthetic image by
stitching local image patches in a kinematically constrained manner. 
The resulting images are used to train an
end-to-end CNN for full-body 3D pose estimation. We cluster the
training data into a 
large number of pose classes and tackle pose estimation as a K-way
classification problem. Such an approach is viable only with 
large training sets such as ours. Our method outperforms
the state of the art in terms of 3D pose estimation in controlled
environments (Human3.6M) and shows promising results for in-the-wild
images (LSP). This demonstrates that CNNs trained on artificial
images generalize well to real images. 


\end{abstract}

\vspace{-3mm}

\section{Introduction}

\vspace{-2mm}



Convolutionnal Neural Networks (CNN) have been very successful for
many different tasks in computer vision. However, training these deep
architectures requires large scale datasets which are not always
available or easily collectable. This is particularly the case for 3D
human pose estimation, for which an accurate annotation of 3D articulated poses in large collections of real images is non-trivial: annotating 2D images with 3D pose information is impractical~\cite{bourdev2009poselets} while large scale 3D pose capture is only available through marker-based systems in constrained environments~\cite{IonescuPOS14}. The images captured in such conditions do not match well real environments. This has limited the development of end-to-end CNN architectures for in-the-wild 3D pose understanding.


Learning architectures usually augment existing training data by
applying synthetic perturbations to the original images,
e.g. jittering exemplars  or applying more complex affine or
perspective transformations \cite{JaderbergSZK15}. Such
data augmentation has proven to be a crucial stage, especially for
training deep architectures.
Recent
work~\cite{JaderbergSVZ16,PengSAS15,SuQLG15,WuSKYZTX15_ShapeNets} has
introduced the use of data synthesis as a solution to train CNNs when
only limited data is available. Synthesis can potentially provide
infinite training data by rendering 3D CAD models from any camera
viewpoint~\cite{PengSAS15,SuQLG15,WuSKYZTX15_ShapeNets}. Fisher et
al~\cite{DosovitskiyFIHH15_FlowNet} generate a synthetic ``Flying
Chairs'' dataset to learn optical flow with a CNN and show that networks
trained on this unrealistic data still generalize very well to
existing datasets. In the context of scene text recognition, Jaderberg
et al.~\cite{JaderbergSVZ16} trained solely on data produced by a
synthetic text generation engine. In this case, the synthetic data is
highly realistic and sufficient to replace real data. Although
synthesis seems like an appealing solution, there often exists a large
domain shift from synthetic to real data~\cite{PengSAS15}. Integrating
a human 3D model in a given background in a realistic way is not
trivial. Rendering a collection of photo-realistic images  (in terms
of color, texture, context, shadow) that would cover the variations in
pose, body shape, clothing and scenes is a challenging task.  

\begin{figure}[htb]
   \centering
  \includegraphics[width=\textwidth]{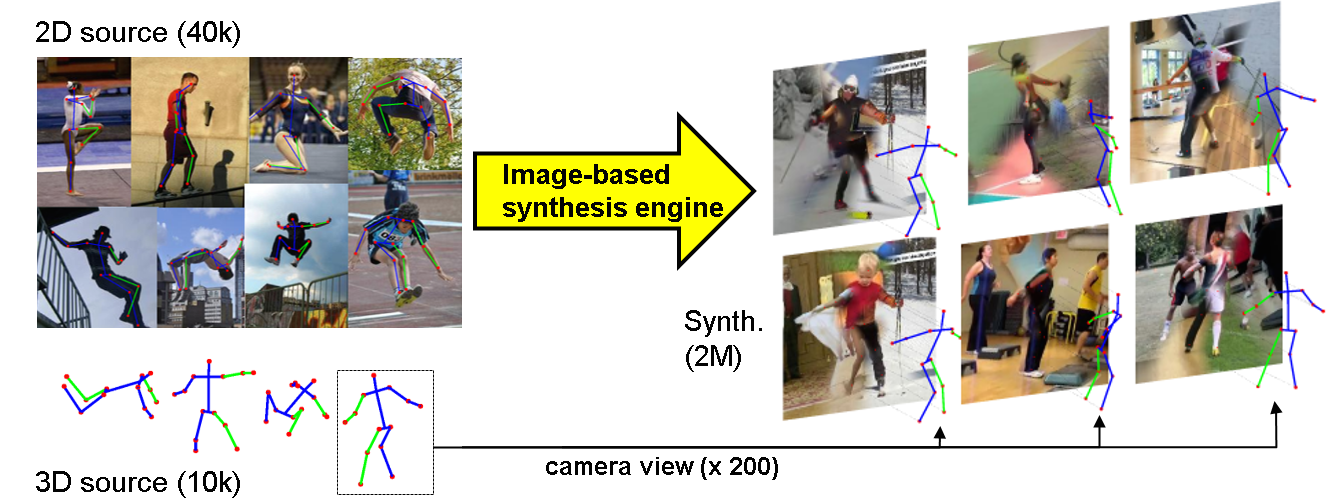}
   \vspace{-7mm}\caption{\footnotesize Image-based synthesis engine. Input: real images with manual annotation of 2D poses, and 3D poses
     captured with a Motion Capture (MoCap) system.  Output:
     220x220 synthetic images and associated 3D poses.
}

     \label{fig:dataset}
 \end{figure}

Instead of rendering a human 3D model, we propose
an image-based synthesis approach that makes use of Motion Capture
(MoCap) data to augment an existing dataset of real images with 2D
pose annotations. Our system synthesizes a very large number of new
in-the-wild images showing more pose configurations and, importantly,
it provides the corresponding 3D pose annotations (see
Fig.~\ref{fig:dataset}).
For each candidate 3D pose in the MoCap library, our system combines
several annotated images to generate a synthetic image of a human in
this particular pose. This is achieved  by ``copy-pasting'' the image
information corresponding to each joint in a kinematically constrained manner.
Given this large ``in-the-wild'' dataset, we implement an end-to-end
CNN architecture for 3D pose estimation. Our approach first clusters
the 3D poses into K pose classes. Then, a K-way CNN
classifier is trained to return a distribution over probable pose
classes given a bounding box around the human in the image. Our method
outperforms state-of-the-art results in terms of 3D pose estimation in
controlled environments and shows promising results on images captured
``in-the-wild''. 
 \vspace{-2mm}
\vspace{-3mm}

\subsection{Related work}



{\bf 3D human pose estimation in monocular images.} 
Recent approaches employ CNNs for 3D pose estimation in monocular
images~\cite{LiZC15} or in videos~\cite{ZhouZLDD16}. Due to the
lack of large scale training data, they are usually trained  (and
tested) on 3D MoCap data in constrained
environments~\cite{LiZC15}. Pose understanding in natural images
is usually limited to 2D pose
estimation~\cite{ChenY14,TompsonJLB14,ToshevS14_DeepPose}. Recent work
also tackles 3D pose understanding from 2D
poses~\cite{AkhterB15,FanZZW14}. Some
approaches use as input the 2D joints automatically provided by a 2D
pose detector~\cite{Simo-SerraRATM12,WangWLYG14}, while others jointly
solve  the  2D and 3D pose estimation~\cite{Simo-SerraQTM13,ZhouT14}.
Most similar to ours is the approach of Iqbal et al.~\cite{IqbalGG16}
who use a dual-source approach that combines 2D pose estimation with
3D pose retrieval. Our method uses the same two training sources,
i.e., images with annotated 2D pose and 3D MoCap data. However, we
combine both sources off-line to generate a large training set that is
used to train an end-to-end CNN 3D pose classifier. This is shown to
improve over~\cite{IqbalGG16}, which can be explained by the fact that
training is performed in an end-to-end fashion.

\noindent {\bf Synthetic pose data.} A number of works have considered the use of synthetic data for human pose estimation. Synthetic data have been used for upper body~\cite{ShakhnarovichVD03}, full-body silhouettes~\cite{DAgarwalT06}, hand-object interactions~\cite{romero_hands_2010}, full-body pose from depth~\cite{ShottonFCSFMKB11} or egocentric RGB-D scenes~\cite{RogezSR15}. Recently,  Zuffi and Black~\cite{ZuffiB15} used a 3D mesh-model to sample synthetic exemplars and fit 3D scans. In~\cite{HattoriBKK15}, a scene-specific pedestrian detectors was learned  without real data while~\cite{EnzweilerG08} synthesized  virtual samples with a generative model to enhance the classification performance of a discriminative model. In~\cite{HornungDK07},  pictures of 2D characters were animated by fitting and deforming a 3D mesh model. Later, \cite{PishchulinJATS12} augmented labelled training images with small perturbations in a similar way. These methods require a perfect segmentation of the humans in the images. Park and Ramanan~\cite{ParkR15}  synthesized hypothetical poses for tracking purposes by applying geometric transformations to the first frame of a video sequence. We also use  image-based synthesis to generate images but our rendering engine combines image regions from several images to create  images with associated 3D poses.  


 \section{Image-based synthesis engine}
  
\begin{figure}[t]
  \centering
   \hspace{-0mm}\includegraphics[width=\textwidth]{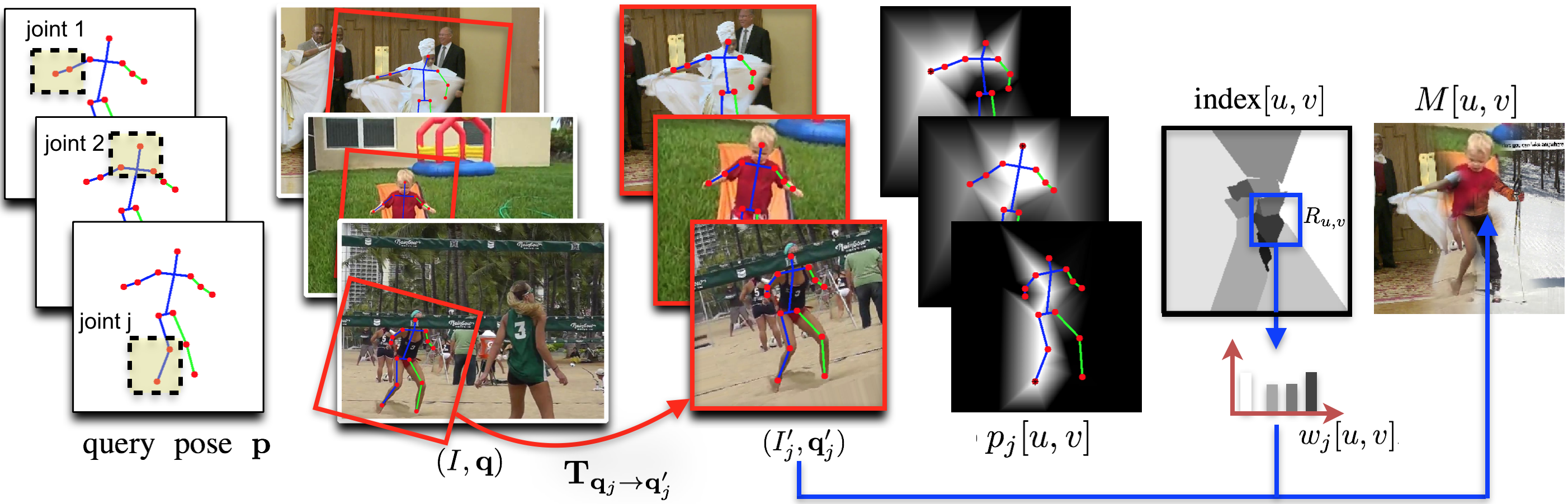}
     \vspace{-5mm} \caption{\footnotesize Synthesis engine. From left to right:  for each joint $j$ of a 2D query pose ${{\bf p}}$ (centered in a  $220 \times 220$ bounding box), we align all the annotated 2D poses w.r.t the limb and search for the best pose match, obtaining a list of $n$ matches $\{(I_j',    {    \bf q}_j'), j=1...n\}$ where $I_j'$ is obtained after transforming $I_j$ with ${{\bf T}{{\bf q}_j \rightarrow {\bf q}_j'}}$. For each retrieved pair, we compute a probability map $p_j[u,v]$. These $n$ maps are used to compute  $\text{index}[u,v] \in \{1...n\}$, pointing to the image $I_{j}'$ that should be used for a particular pixel $(u,v)$. Finally, our blending algorithm computes each pixel value of the synthetic image $M[u,v]$ as the weighted sum over all aligned images $I_j'$, the weights being calculated using an histogram of indexes in a squared region $R_{u,v}$ around  $(u,v)$.}
      \label{fig:mosaique}
\end{figure}

 At the heart of our approach is an image-based synthesis engine that
 artificially generates ``in-the-wild'' images  with 3D pose
 annotations. Our method takes as input a dataset of real images with 2D
 annotations and a library of 3D Motion Capture (MoCap)
 data, and generates a large number of synthetic images with associated
 3D poses (Fig.~\ref{fig:dataset}). 
We introduce an image-based rendering engine that augments the
existing database of annotated images with a very large set of
photorealistic images covering more body pose configurations
than the original set. This is done by selecting and stitching image
patches in a kinematically constrained manner using the MoCap 3D
poses. 
Our synthesis process consists
 of two stages:  a MoCap-guided mosaic construction stage that stitches image
 patches together and a pose-aware blending  process that improves image quality and erases  patch seams. These are discussed in the following subsections. 
 Fig.~\ref{fig:mosaique} summarizes the overall process. 


 

 \subsection{MoCap-guided image mosaicing}

Given a 3D pose with $n$ joints  ${\bf P \in \mathbb{R}^{n \times
    3}}$, and its projected 2D joints ${{\bf p}=\{p_{j} , j=1...n\}}$  in  a particular camera view, we want to find for each joint $j \in \{1...n \}$ an image whose annotated 2D pose presents a similar kinematic configuration around $j$. To do so, we define a distance function between 2 different 2D poses ${\bf p}$ and ${\bf q}$, conditioned on joint $j$ as:
\begin{align}
  D_{j}({\bf p}, {\bf q})= \sum_{k=1}^{n} d_{\text{E}}(p_k, q'_k)  \label{eq:posedist} 
\end{align}
where $d_{\text{E}}$ is the Euclidean distance. ${\bf q'}$ is the
aligned version of ${\bf q}$ with respect to joint $j$ after applying
a rigid transformation ${{\bf T}_{{\bf q}_j \rightarrow {\bf q}_j'}}$,
which respects $q_{j}'=p_{j}$ and $q_{i}'=p_{i}$ , where $i$ is the
farthest directly connected joint to $j$ in ${\bf p}$. This function $D_{j}$ measures the similarity between 2 joints by aligning and taking into account the entire poses. 
To increase the influence of neighboring joints, we weight the distances $d_{\text{E}}$ between each pair of joints $\{(p_k, q'_k), k=1...n\}$  according to their distance to the query joint $j$ in both poses. Eq.~\ref{eq:posedist} becomes:
 \begin{align}
   D_{j}({\bf p}, {\bf q})= \sum_{k=1}^{n} (w_k^j({\bf p}) + w_k^j({\bf q})) \;d_{\text{E}}(p_k, q'_k) \label{eq:posesim} 
 \end{align}
where weight $w_k^j$ is inversely proportional to the distance between joint $k$ and the query joint $j$, i.e., $w_k^j({\bf p})=1/d_{\text{E}}(p_k, p_j)$ and normalized so that $\sum_k w_k^j({\bf p})=1$. For each joint $j$ of the query pose  ${{\bf
    p}}$,
we retrieve from our dataset  $\mathbb{Q}=\{(I_1,{\bf q}_1)\dots(I_N,{\bf q}_N)\}$ of images and annotated 2D poses\footnote{In practice, we do not search for occluded joints.}:
 \begin{align}
  {\bf q}_j = \text{argmin}_{{\bf q} \in \mathbb{Q}} D_{j}({\bf p}, {\bf q})  \quad \forall j  \;  \in \{1...n\}. \label{eq:posesearch} 
 \end{align}
We obtain a list of $n$ matches $\{(I_j',{\bf q}_j'), j=1...n\}$ where
$I_j'$ is the cropped image obtained after transforming $I_j$ with
${{\bf T}_{{\bf q}_j \rightarrow {\bf q}_j'}}$. Note that a same pair
$(I,{\bf q})$ can appear multiple times in the list of candidates,
i.e., being a good match for several joints. 

Finally, to render a new image, we need to select the candidate images $I_j'$ to be used for each pixel $(u,v)$. Instead of using regular patches, we compute a probability map $p_j[u,v]$ associated with each pair $(I_j',{\bf q}_j')$ based on local matches measured by $d_{\text{E}}(p_k, q'_k)$ in Eq.~\ref{eq:posedist}. To do so, we first apply a Delaunay triangulation to the set of 2D joints in $\{{\bf q}_j'\}$  obtaining a partition of the image into triangles, accordingly to the selected pose. Then, we assign the probability $p_j(q'_k)=exp(-d_{\text{E}}(p_k, q'_k)^2/\sigma^2) $ to each vertex $q'_k$. We finally compute a probability map $p_j[u,v]$ by interpolating values from these vertices using barycentric interpolation inside each triangle.
The resulting $n$ probability maps are concatenated and  an index map $\text{index}[u,v] \in \{1...n\}$ can be computed as follows: 
\begin{align}
\text{index}[u,v]=\text{argmax}_{j \in \{1 \dots n\}} \; p_j[u,v], 
\end{align}
this map pointing to the training image $I_{j}'$ that should be used for each pixel $(u,v)$. A mosaic $M[u,v]$ can be generated by ``copy-pasting'' image information at pixel $(u,v)$ indicated by $\text{index}[u,v]$:
\begin{align}
M[u,v]=I_{j^*}'[u,v] \quad \text{with} \quad j^*=\text{index}[u,v]. 
\end{align}

 \subsection{Pose-aware image blending}
 
The mosaic $M[u,v]$ resulting from the previous stage presents significant artifacts at the boundaries between image regions. Smoothing is necessary to prevent the learning algorithm from interpreting these artifacts as discriminative pose-related features. We first experimented with off-the-shelf image filtering and alpha blending algorithms, but the results were not satisfactory. Instead, we propose a new pose-aware blending algorithm that maintains image information on the human body while erasing most of the stitching artifacts. 
For each pixel $(u,v)$, we select a surrounding squared region $R_{u,v}$ whose size varies with the distance of pixel $(u,v)$ to the pose: $R_{u,v}$ will be larger when far from the body and smaller nearby. Then, we evaluate how much each image $I_{j}'$ should contribute to the value of pixel $(u,v)$ by building an histogram of the image indexes inside the region $R_{u,v}$:
 \begin{align}
   w_j[u,v]=\text{Hist}(\text{index}(R_{u,v})) \; \forall j \in \{ 1 \dots n \},
 \end{align}
 where the weights are normalized so that $\sum_j w_j[u,v] =1$. The final mosaic $M[u,v]$ (see examples in Fig.~\ref{fig:dataset}) is then computed as the weighted sum over all aligned images:
 \begin{align}
   M[u,v] = \sum_j w_j[u,v] I_j'[u,v].
 \end{align}
This procedure produces plausible images that are kinematically correct and locally photorealistic. 
 
 \section{CNN for full-body 3D pose estimation}
 


\begin{figure}[t]
   \centering 
  \hspace{-0mm}\includegraphics[width=0.9\textwidth]{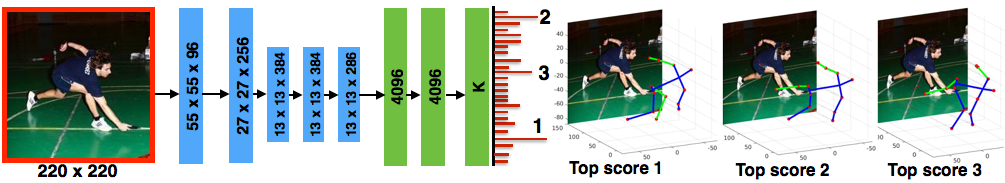}
   \vspace{-3mm}\caption{\footnotesize  CNN-based pose classifier. We show the different layers with their corresponding dimensions, with convolutional layers depicted in blue and fully connected ones in green.  
The output is a distribution over K pose classes. Pose estimation is obtained by taking the highest score in this distribution. We show on the right the 3D poses for 3 highest scores.}
     \label{fig:CNN}
 \end{figure}

Human pose estimation has been addressed as a classification problem in the past~\cite{BissaccoYS06,OkadaS08, RogezSR15, RogezROT12}. Here, the 3D pose space is partitioned into K clusters and a K-way classifier is trained to return a distribution over pose classes. Such a classification approach allows modeling multimodal outputs in ambiguous cases, and produces multiple hypothesis that can be rescored, e.g.\ using temporal information. Training such a classifier requires a reasonable amount of data per class which implies a well-defined and limited pose space (e.g. walking action)~\cite{RogezROT12,BissaccoYS06}, a large-scale synthetic dataset~\cite{RogezSR15} or both~\cite{OkadaS08}.  
Here, we introduce a CNN-based classification approach for full-body 3D pose estimation. Inspired by the DeepPose algorithm~\cite{ToshevS14_DeepPose} where the AlexNet CNN architecture \cite{NIPS2012_Alexnet} is used for full-body 2D pose regression, we select the same architecture and adapt it to the task of 3D body pose classification. This is done by adapting the last fully-connected layer to output a distribution of scores over pose classes as illustrated in Fig.~\ref{fig:CNN}. 
 Training such a classifier requires a large amount of training data that we generate using our image-based synthesis engine. 

Given a library of MoCap data and a set of camera views, we synthesize for each 3D pose a $220\times220$ image
. This size has proved to be adequate for full-body pose estimation~\cite{ToshevS14_DeepPose}. The 3D poses are then aligned with respect to the camera center and translated to the center of the torso. 
In that way, we obtain orientated 3D poses that also contain the viewpoint information. 
 We cluster the resulting 3D poses to define our classes which will correspond to groups of similar orientated 3D poses.
 We empirically found that K=5000 clusters was a sufficient number of clusters
 . For evaluation, we return the average 2D and 3D poses of the top scoring class. 

To compare with~\cite{ToshevS14_DeepPose}, we also train a holistic pose regressor, which regresses to 2D and 3D poses (not only 2D). To do so, we concatenate the 3D coordinates expressed in meters normalized to the range $[-1,1]$,  with the 2D pose coordinates, also normalized in the range $[-1,1]$ following~\cite{ToshevS14_DeepPose}.


 

  

 \section{Experiments}

We address 3D pose estimation in the wild. However, there does not
exist a dataset of real-world images with 3D annotations. We thus
evaluate our method in two different settings using existing datasets:
(1) we validate our 3D pose predictions using 
Human3.6M~\cite{IonescuPOS14} which provides accurate 3D and 2D poses for 15
different actions captured in a controlled indoor
environment; (2) we evaluate on Leeds Sport dataset
(LSP)\cite{JohnsonE10} that presents in-the-wild images together with
full-body 2D pose annotations.  We demonstrate competitive results
with state-of-the-art methods for both of them. 

Our image-based rendering engine requires two different training
sources: 1) a 2D source of images with 2D pose annotations and 2) a MoCap 3D source. We consider two different datasets for each: for 3D
poses we use the CMU Motion Capture
Dataset\footnote{http://mocap.cs.cmu.edu} and the Human3.6M 3D
poses~\cite{IonescuPOS14}, and for 2D pose annotations the
MPII-LSP-extended dataset~\cite{PishchulinITAAG15} and the Human3.6M 2D poses and images.




{\noindent \bf MoCap 3D source.} The CMU Motion Capture dataset consists of 2500 sequences and a total of 140,000 3D poses. We align the 3D poses w.r.t. the torso and select a subset of 12,000 poses, ensuring that selected poses have at least one joint 5 cm apart. In that way, we densely populate our pose space and avoid repeating common poses (e.g. neutral standing or walking poses which are over-represented in the dataset). For each of the 12,000 original MoCap poses, we sample 180 random virtual views with azimuth angle spanning 360 degrees and elevation angles in the range $[-45,45]$. We generate over 2 million pairs of 3D/2D pose configurations (articulated poses + camera position and angle). 
For Human3.6M, we 
randomly selected a subset of 190,000 orientated 3D poses, discarding similar poses, i.e., when the average Euclidean distance of the joints is less than 15mm as in~\cite{IqbalGG16}.

{\noindent \bf 2D source.}  For the training dataset of real
images with 2D pose annotations, we use the MPII-LSP-extended
\cite{PishchulinITAAG15} which is a concatenation of the extended LSP
\cite{JohnsonE11} and the MPII dataset \cite{andriluka14cvpr}. Some of
the poses were manually corrected as a non-negligible number of
annotations are not accurate enough or completely wrong (eg., right-left inversions or bad ordering of the joints along a limb). 
 We mirror the images to double the size of the training set,
 obtaining a total of 80,000 images with 2D pose annotations.
For Human3.6M, we consider 
the 4 cameras and  create  a pool of 17,000
images and associated 2D poses that we also mirror. We ensure that
most similar poses have at least one joint 5 cm apart in 3D.

 \subsection{Evaluation on Human3.6M Dataset (H3.6M)}

To compare our results with very recent work in 3D pose estimation~\cite{IqbalGG16}, we follow the protocol introduced in \cite{KostrikovG14} and employed in \cite{IqbalGG16}: we consider six subjects (S1, S5, S6, S7, S8 and S9) for training, use every $64^{th}$ frame of subject S11 for testing and evaluate the 3D pose error (mm) averaged over the 13 joints. We refer to this protocol by P1. As in~\cite{IqbalGG16}, we consider a 3D pose error that measures accuracy of aligned  pose by a rigid transformation but also report the absolute error.  

We first evaluate the impact of our synthetic data on the performances
for both the regressor and classifier. The results are reported in
Tab.~\ref{tab:H36M}. We can observe that when considering few training
images (17,000), the regressor clearly outperforms the classifier
which, in turns, reaches better performances when trained on larger
sets. This can be explained by  
the fact that the classification approach requires a sufficient amount of examples.
 We, then, compare results when training both regressor and classifier
 on the same 190,000 poses considering a) synthetic data generating
 from H3.6M, b) the real images corresponding to the 190,000 poses and
 c) the synthetic and real images together. We observe that the
 classifier has similar performance when trained on synthetic or
 real images, which means that our image-based rendering engine
 synthesizes useful data. Furthermore, we can see that the classifier
 performs much better  when trained on synthetic and real images
 together. This means that our data  is different from
 the original data and allows the classifier to learn better
 features. Note that we retrain Alexnet from scratch. We found that it performed better than just fine-tuning a model pre-trained on Imagenet (3D error of 88.1mm vs 98.3mm with fine-tuning).
  \begin{table} 
\caption{3D pose estimation results on Human3.6M (protocol P1).}
\vspace{-2mm}\centering
\begin{tabular}{c|c|c|c|c }
   Method & Type  of images & 2D source size    & 3D source size & Error (mm)  \\  
\hline 
Reg. & Real &  17,000 &  17,000  & 112.9   \\ 
Class. & Real &    17,000 &  17,000  &149.7   \\ 
\hline
Reg. & Synth &  17,000 & 190,000  & 101.9 \\ 
Class. & Synth &  17,000   & 190,000 &  97.2    \\ 
\hline
Reg. & Real & 190,000   & 190,000  &139.6    \\ 
Class. & Real & 190,000   & 190,000  &97.7  \\ 
\hline
Reg. & Synth + Real &  207,000 & 190,000 &125.5  \\ 
Class. & Synth + Real & 207,000  & 190,000 &\bf{88.1}   \\ 
\end{tabular} 
\label{tab:H36M}
\end{table}

In Tab.~\ref{tab:H36M_sota}, we compare our results to
  state-of-the-art approaches. We also report results for a second
  protocol (P2) employed in~\cite{LiZC15,ZhouZLDD16,TekinRLF16} where
  all the frames from subjects S9 and S11 are used for testing and
  only S1, S5, S6, S7 and S8 are used for training. Our best
  classifier, trained with a combination of synthetic and real data,
  outperforms state-of-the-art results in terms of 3D pose estimation
  for single frames. Zhou et al.~\cite{ZhouZLDD16} report better
  performance, but they integrate temporal information. 
Note that our method estimates absolute pose (including orientation
w.r.t. the camera), which is not the case for other methods such as Bo
et al.~\cite{BoS10}, who estimate a relative pose and do not provide
3D orientation.



  \begin{table} 
\caption{Comparison with state-of-the-art results on Human3.6M. The average 3D pose error (mm) is reported before (Abs.) and after rigid 3D alignment for 2 different protocols. See text for details.}
\vspace{-2mm}\centering
\begin{tabular}{c|c|c|c|c}
  Method &  Abs. Error (P1)  &  Error (P1) &  Abs. Error (P2)  &  Error (P2) \\ 
\hline
Bo\&Sminchisescu~\cite{BoS10}  & - & 117.9 & - & -\\
Kostrikov\&Gall~\cite{KostrikovG14} &  - &115.7 & -& -\\
Iqbal et al.~\cite{IqbalGG16} & - & 108.3 & -& -\\ 
\hline  
Li et al.~\cite{LiZC15} & - & - &  121.31 & -\\ 
Tekin et al.~\cite{TekinRLF16}& - & - &  124.97 & -\\ 
Zhou et al.~\cite{ZhouZLDD16}& - & - &  \bf{113.01} & -\\ 
\hline  
Ours  &126 & \bf{88.1} & 121.2 & 87.3\\
\end{tabular} 
\label{tab:H36M_sota}
\end{table}

 \subsection{Evaluation on Leeds Sport Dataset (LSP)}

We now train our pose classifier using different combinations of
training sources and use them to estimate 3D poses on images captured
in-the-wild, i.e., LSP. Since 3D pose evaluation is not possible on
this dataset, we instead compare 2D pose errors expressed in
pixels and measure this error on the normalized $220 \times 220$ images following~\cite{ZhouZLDD16}. We compute the average 2D
pose error over the 13 joints on both LSP and H3.6M (see
Table~\ref{tab:LSPpix}).

\begin{figure}[htb]
   \centering 
  \hspace{-0mm}\includegraphics[width=\textwidth]{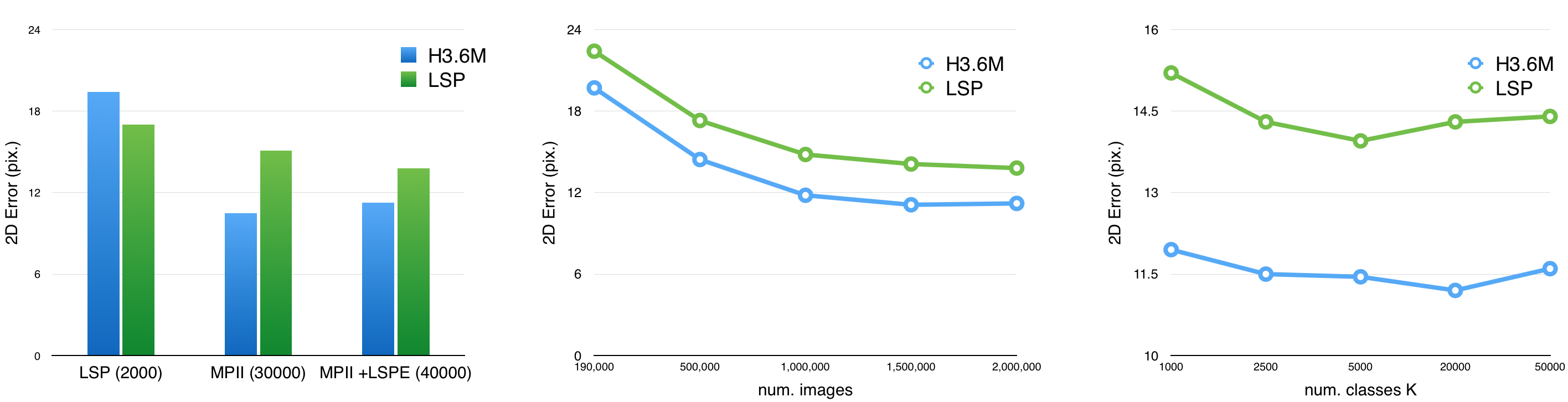}
\vspace{-5mm}  \caption{\footnotesize  2D pose error on LSP and
  Human3.6M using different pools of annotated images to generate 2
  million of synthetic training images (left), varying the number of
  synthetic training images (center) and considering different number of pose classes K (right).}
     \label{fig:plots}
 \end{figure}

As expected, we observe that when using a pool of the in-the-wild
images to generate the synthetic data, the performance
increases on LSP and drops on H3.6M, showing the importance of
realistic images for good performance in-the-wild and the lack of
generability of models trained on constrained indoor images. The  error slightly
increases in both cases when using the same number (190,000) of CMU 3D
poses. The same drop was observed by~\cite{IqbalGG16} and can be
explained by the fact that by CMU data covers a larger portions of the
3D pose space, resulting in a worse fit. The results improve on both
test sets when considering more poses and synthetic images (2
millions). The larger drop in Abs 3D error and 2D error compared to 3D error means that a better 
camera view is estimated when using more synthetic data.
In all cases, the  performance (in pixel) is lower on LSP than on
H3.6M due to the fact that the poses observed in LSP are more
different from the ones in the CMU MoCap data. In Fig.~\ref{fig:plots}
, we visualize the 2D pose error on LSP and Human3.6M 1) for different
pools of annotated 2D images, 2) varying the number of synthesized
training images and 3) considering  different number of pose classes
K. As expected using a bigger set of annotated images improves the
performance in-the-wild. Pose error converges both on LSP and H3.6M
when using 1.5 million of images; using more than $K=5000$ classes does not further improve the performance.

  \begin{table} 
\caption{Pose error on LSP and H3.6M using different sources for rendering the synthetic images.}
\vspace{-2mm}\centering
\begin{tabular}{c|c|c|c|c|c|c}
    2D   & 3D   & Num. of & H3.6M & H3.6M   & H3.6M   & LSP \\ 
       source &   source & 3D poses& Abs Error (mm)&   Error (mm) &    Error (pix) &   Error (pix)\\ 
\hline 
 H3.6M   & H3.6M & 190,000 & 130.1&  97.2 &8.8&31.1 \\  
 MPII+LSP   & H3.6M &190,000 &248.9& 122.1 &17.3&20.7   \\ 
 MPII+LSP     & CMU & 190,000 &  320.0&150.6 &19.7&22.4  \\ 
 MPII+LSP   & CMU &  $2.10^6$   & 216.5 & 138.0 & 11.2 & 13.8  \\ 
\end{tabular} 
\label{tab:LSPpix}
\end{table}


 \begin{figure}[t]
\vspace{-2mm}
   \centering 
  \hspace{-0mm}\includegraphics[width=\textwidth]{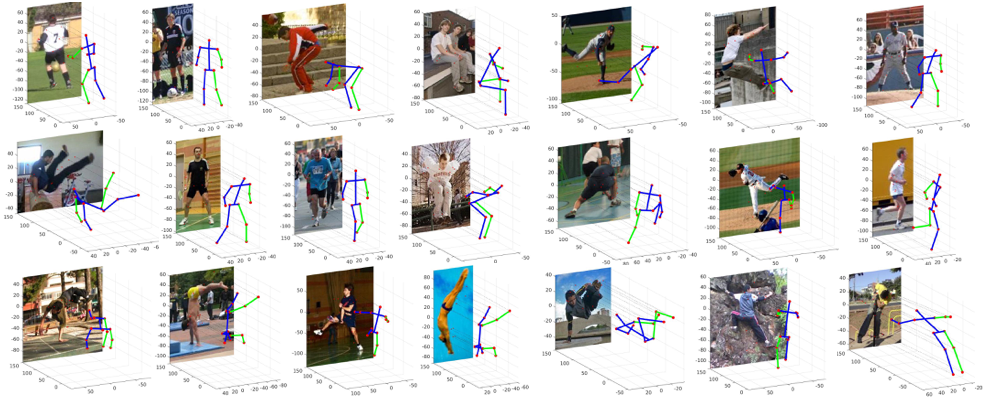}
  \vspace{-3mm} \caption{\footnotesize Qualitative results on LSP. 
We show correct 3D pose estimations (top 2 rows) and typical failure cases (bottom row) corresponding to unseen poses or right-left and front-back confusions.}
 
     \label{fig:results}
 \end{figure}

To further improve the performance, we also experiment with fine-tuning a VGG-16 architecture \cite{SimonyanZ14a} for pose classification. By doing so, the average (normalized) 2D pose error decreases by $2.3$ pixels.
In Table~\ref{tab:LSPsota}, we compare our results on LSP to the
state-of-the-art 2D pose estimation methods. Although our
approach is designed to estimate a coarse 3D pose, its performances is
comparable to recent 2D pose
estimation methods~\cite{ChenY14,yang2016end}. 
 
The qualitative results in Fig.~\ref{fig:results} show that our algorithm correctly estimates the global 3D pose. After a visual analysis of the results, we found that failures occur in two cases:  1) when the observed pose does not belong to the MoCap training database, which is a limitation of purely holistic  approaches, or 2) when there is a possible right-left or front-back confusion. We observed that this later case is often correct for subsequent top-scoring poses.  This highlights a property of our approach that can keep multiple pose hypotheses which could be rescored adequately, for instance, using temporal information in videos. 
  \begin{table} 
\caption{State-of-the-art results on LSP (2D pose error in pixels on normalized $220 \times 220$ images).}
\vspace{-2mm}\centering
\begin{tabular}{c|c|c|c|c|c|c|c|c}
    Method   & Feet   & Knees & Hips & Hands  & Elbows & Shoulder & Head & All\\ 
\hline  

Wei et al.~\cite{WeiRKS16} &6.6 & 5.3& 4.8& 8.6 & 7.0& 5.2& 5.3& {\bf6.2}\\
Pishchulin et al.~\cite{PishchulinITAAG15} & 10.0 & 6.8 & 5.0 &  11.1 & 8.2 & 5.7 & 5.9 & 7.6\\   
Chen \& Yuille  ~\cite{ChenY14} &15.7 & 11.5& 8.1&  15.6 & 12.1& 8.6  & 6.8 & 11.5\\   
Yang et al.~\cite{yang2016end} &15.5 & 11.5& 8.0 &  14.7 & 12.2& 8.9 & 7.4& 11.5\\
   \hline
Ours (Alexnet) & 19.1 & 13 & 4.9 &  21.4 & 16.6& 10.5 & 10.3& 13.8  \\ 
Ours (VGG)  & 16.2 & 10.6 & 4.1 &  17.7 & 13.0& 8.4 &9.8& 11.5\\  
\end{tabular} 
\label{tab:LSPsota}
\end{table}

 \section{Conclusion}

In this paper, we introduce an approach for creating a synthetic training dataset of  ``in-the-wild'' images and their corresponding 3D pose.  
Our algorithm artificially augments a dataset of real images with new synthetic
images showing new poses and, importantly, with 3D pose annotations. 
We show that CNNs can be trained on artificial images and
  generalize well to real images. We train an end-to-end CNN classifier for 3D pose estimation and 
show that, with our synthetic training images, our method
outperforms state-of-the-art results in terms of 3D pose estimation in
controlled environments and shows promising results for in-the-wild images (LSP). 
In this paper, we have estimated a coarse 3D pose by returning the
average pose of the top scoring cluster. In future work, we will
investigate how top scoring classes could be re-ranked and also how
the pose could be refined.


\paragraph{\bf Acknowledgments.}This work was supported by the European Commission under FP7 Marie Curie IOF grant (PIOF-GA-2012-328288) and partially supported by ERC advanced  grant Allegro.
We acknowledge the support of NVIDIA with the donation of the GPUs
used for this research. 
We thank P. Weinzaepfel for his help and the anonymous reviewers for their comments and suggestions.


 \footnotesize
\bibliographystyle{plain}
\bibliography{greg_bib}

\end{document}